 \documentclass[pmlr,twocolumn,10pt]{jmlr} 

\usepackage{booktabs}
\usepackage[load-configurations=version-1]{siunitx} 

\theorembodyfont{\upshape}
\theoremheaderfont{\scshape}
\theorempostheader{:}
\theoremsep{\newline}

\jmlrvolume{LEAVE UNSET}
\jmlryear{2021}
\jmlrsubmitted{LEAVE UNSET}
\jmlrpublished{LEAVE UNSET}
\jmlrworkshop{Machine Learning for Health (ML4H) 2021} 

\title[Deconfounding Temporal Autoencoder]{Deconfounding Temporal Autoencoder: Estimating Treatment Effects over Time Using Noisy Proxies}

\author{%
\Name{First Author 1}\Email{abc@sample.com}\\
\addr University X, Country 1
\AND
\Name{First Author 2} \Email{def@sample.com}\\
\addr University Y, Country 2
\AND
\Name{Last Author} \Email{ghi@sample.com}\\
\addr University Z, Country 3
}

\author{
\Name{Milan Kuzmanovic} \Email{mkuzmanovic@ethz.ch}\\
\addr ETH Zurich
\AND
\Name{Tobias Hatt} \Email{thatt@ethz.ch}\\
\addr ETH Zurich
\AND
\Name{Stefan Feuerriegel} \Email{feuerriegel@lmu.de}\\
\addr ETH Zurich, LMU Munich
}

\usepackage{xspace}
\usepackage{algorithmic}

\newcommand{\DTA}{DTA\xspace}
\newcommand{\longname}{\emph{Deconfounding Temporal Autoencoder}\xspace}

\newcommand\ie{i.\,e.}
\newcommand\eg{e.\,g.}

\begin{document}

\maketitle

\begin{abstract}
Estimating individualized treatment effects (ITEs) from observational data is crucial for decision-making. In order to obtain unbiased ITE estimates, a common assumption is that all confounders are observed. However, in practice, it is unlikely that we observe these confounders directly. Instead, we often observe noisy measurements of true confounders, which can serve as valid proxies. In this paper, we address the problem of estimating ITE in the longitudinal setting where we observe noisy proxies instead of true confounders. To this end, we develop the \longname (\DTA), a novel method that leverages observed noisy proxies to learn a hidden embedding that reflects the true hidden confounders. In particular, the \DTA combines a long short-term memory autoencoder with a causal regularization penalty that renders the potential outcomes and treatment assignment conditionally independent given the learned hidden embedding. Once the hidden embedding is learned via \DTA, state-of-the-art outcome models can be used to control for it and obtain unbiased estimates of ITE. Using synthetic and real-world medical data, we demonstrate the effectiveness of our \DTA by improving over state-of-the-art benchmarks by a substantial margin.
\end{abstract}


\section{Introduction}


Individualized treatment effect (ITE) estimation is an important problem for personalized decision-making across many disciplines \citep{Glass2013, Song2019, hatt2020early}. For instance, in medicine, individual treatment effects guide personalized treatment assignments such that the benefit for each patient is maximized. Traditionally, randomized controlled trials (RCTs) represent the gold standard for estimating treatment effects; yet, RCTs are often unfeasible. Consequently, there is a growing need for estimating ITEs from observational data, such as electronic health records.


Numerous methods have been proposed for estimating ITE from observational data in the longitudinal setting \citep{Robins2008, Schulam2017, Lim2018, Bica2020crn}. However, most of these methods assume that all confounders (\ie, variables that affect both treatment assignment and potential outcomes) are observed. When some confounders are unobserved (\ie, hidden), they cannot be controlled for, and, hence, these methods yield ITE estimates that are biased \citep{Pearl2009}. This can lead to sub-optimal or even harmful treatment recommendations. In practice, it is unlikely that we observe all confounders directly. 


In practice, we often find that measurements of confounders are either indirect measurements and/or subject to noise.  Nevertheless, such indirect and/or noisy measurements of true confounders can serve as proxies for the true confounders. For instance, we may not be able to measure socio-economic status of patients directly; however, we can obtain information on their socio-economic status indirectly, such as, \eg, through a patient's job, salary, or residence address. Such information represents proxies of the true socio-economic status. Hence, while the true confounders are often unobserved in practice, observational data usually contains information on the true confounders in form of noisy proxies. Motivated by this, we study how to leverage noisy proxies for unbiased ITE estimation in the longitudinal setting (and thus circumvent issues due to hidden confounders).


In this paper, we propose the \longname (\DTA), a novel method for ITE estimation in the longitudinal setting using noisy proxies. Specifically, \DTA leverages observed noisy proxies to learn a hidden embedding that reflects the true hidden confounders. As such, our \DTA method does not rely on the 'no hidden confounders' assumption; instead, \DTA learns a hidden embedding that reflects the true hidden confounders using noisy proxies in the observational data. The \DTA proceeds in two steps: (i)~we build upon a long short-term memory (LSTM) autoencoder to learn a hidden embedding that leverages information from observed noisy proxies, and (ii)~we add a causal regularization penalty based on the Kullback-Leibler divergence to render potential outcomes and treatment assignment conditionally independent given the hidden embedding. This way, we ensure that the learned hidden embedding reflects the true hidden confounders in terms that it can be used to control for them during ITE estimation. 


We emphasize that our \DTA works as a deconfounding method. As such, it captures information on the true hidden confounders by leveraging the observed noisy proxies and, therefore, enables unbiased ITE estimation in the longitudinal setting. The output of the \DTA is a hidden embedding that reflects the true hidden confounders, and which can as such be used in conjunction with any outcome model for ITE estimation. We demonstrate the effectiveness of our \DTA using experiments with both synthetic and real-world data. First, we perform a simulation study with synthetic data where we control the amount of hidden confounding. Here, we show that, across different amounts of confounding, \DTA is effective in reducing bias in ITE estimation. Second, we use real-world data from patients in the intensive care units \citep{Johnson2016}. Here, our \DTA yields significant performance improvements over state-of-the-art outcome models for ITE estimation in the longitudinal setting. 


We list our main \textbf{contributions}\footnote{Code available at: \url{https://github.com/mkuzma96/DTA}} as follows:
\begin{enumerate}
\item We develop the \longname, a novel method that leverages \emph{noisy proxies} of hidden confounders for ITE estimation in the longitudinal setting. 
\item We propose a causal regularization penalty that forces potential outcomes and treatment assignment to be conditionally independent given the learned hidden embedding.
\vspace{-.35em}
\item We demonstrate that our \DTA, in conjunction with an outcome model, substantially improves over state-of-the-art benchmarks using both synthetic and real-world data.
\end{enumerate}

\section{Related work}
\label{sec:lit}
Extensive work focuses on estimating treatment effects
in the static setting \citep[\eg,][]{Johansson2016, Shalit2017, curth2021inductive, Hatt2021b}. In contrast, our work focuses on ITE estimation in the longitudinal setting where, additionally, we observe noisy proxies instead of the true hidden confounders. Hence, we review two streams in the literature that are particularly relevant, namely (i)~methods for ITE estimation in the longitudinal setting that rely on sequential strong ignorability and (ii)~methods that try to control for the hidden confounders. Additional related work is reviewed in \appendixref{app:relwork}.

\textbf{(i)~ITE estimation in the longitudinal setting.} Many methods for estimating treatment effects in the longitudinal setting originate from the epidemiology literature. Examples include the g-computation formula, g-estimation of structural nested mean models, and inverse probability of treatment weighting in marginal structural models (MSMs) \citep{Robins2000, Robins2008, Hernan2020}. However, these methods rely on linear predictors for estimation and are thus unable to handle more complex disease dynamics. \citet{Lim2018} address this limitation through the use of recurrent marginal structural networks (RMSNs), where nonlinear dependencies are explicitly captured. Additionally, other methods were proposed based on Bayesian nonparametrics \citep{Xu2016, Roy2017} and domain adversarial training \citep{Bica2020crn, berrevoets2020organite}. 

All of the above methods rely on the 'no hidden confounders' assumption, and, therefore, they fail when the true confounders are not directly observed. In contrast, our \DTA aims at learning a hidden embedding that reflects the true hidden confounders from noisy proxies in the observed data and, thus, allow for unbiased ITE estimation in the longitudinal setting with any of the above outcome models. 

\textbf{(ii)~ITE estimation with hidden confounders.} Little work has been done on ITE estimation in the presence of hidden confounders. In the static setting, methods have been proposed to learn substitutes for the hidden confounders from noisy proxies in the observed data, which can then be used for deconfounding. These include methods based on variational autoencoders \citep{Louizos2017} and matrix factorization \citep{Kallus2018}. However, these methods were specifically tailored to the static setting and, hence, are not applicable in the longitudinal setting. Furthermore, \citet{Wang2019} developed a deconfounder method that learns substitutes for hidden confounders using factor models for multiple treatments, but again in the static setting. 

In the longitudinal setting, the ideas from \citet{Wang2019} were extended for deconfounding over time using a recurrent neural network architecture \citep{Bica2020tsd}. In a similar vein, \citet{Hatt2021} proposed a sequential deconfounder for longitudinal setting that infers substitutes for the hidden confounders using Gaussian process latent variable model. However, both deconfounding methods for the longitudinal setting rely on associations between treatments and assume that the observed covariates are true confounders. As such, these methods cannot deal with our setting where observed covariates are noisy proxies of the true confounders. 

To the best of our knowledge, we are the first to consider noisy proxies to learn a hidden embedding that reflects the true hidden confounders and, based on it, estimate ITE in the longitudinal setting.

\section{Problem setup}

\subsection{Setting}

\begin{figure}[tbp]
\centerline{\includegraphics[width=3.9cm]{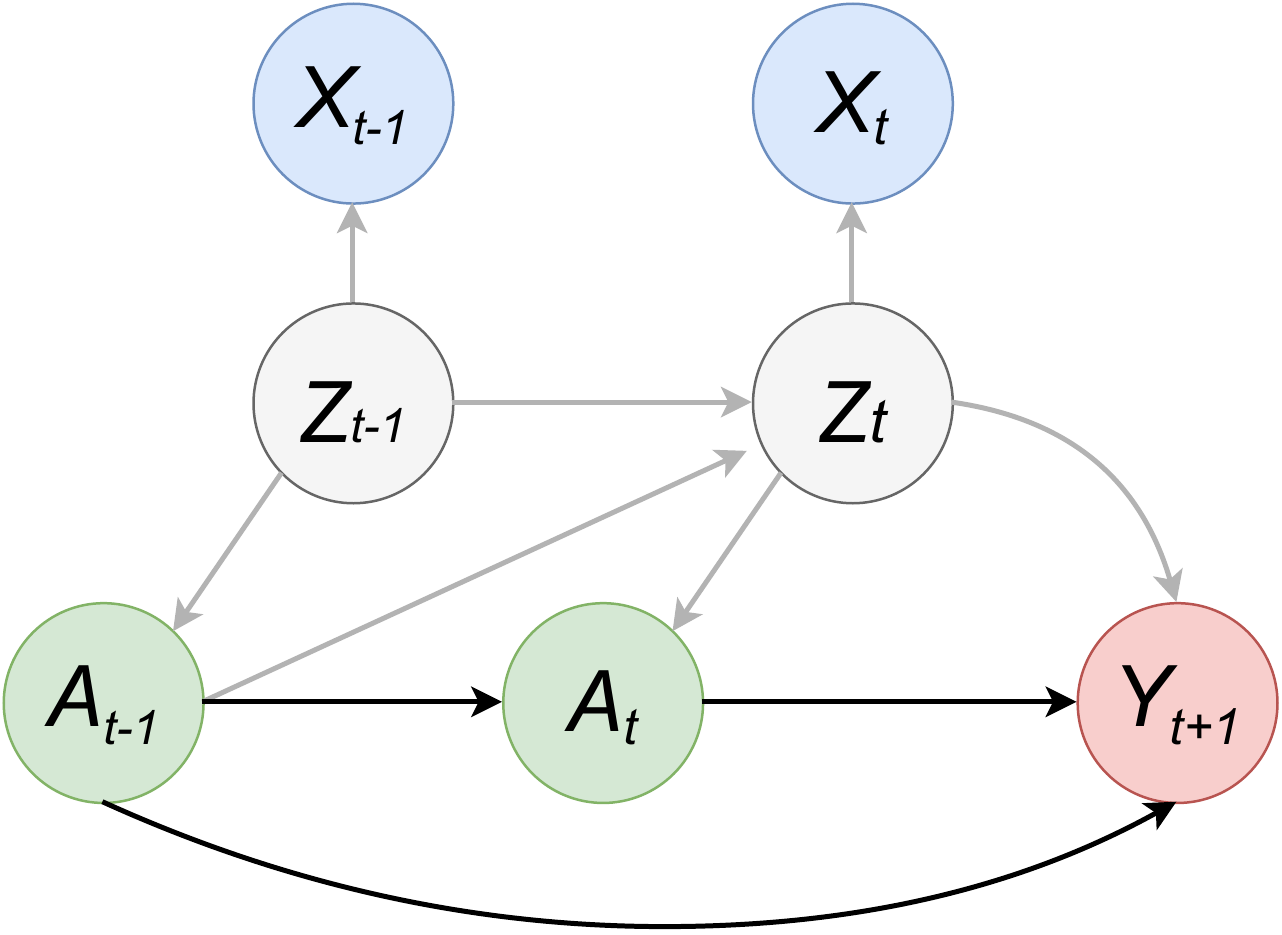}}
\vspace{-.75em}\caption{An overview of the causal structure.}
\label{fig:causal_structure}\vspace{-1.25em}
\end{figure}

We consider a cross-sectional setting with patients $i = 1, \ldots, N$ and timesteps $t = 1, \ldots, T$. At time $t$, let the random variables $\mathbf{Z}_t^{(i)} = [ Z_{t,1}^{(i)}, Z_{t,2}^{(i)}, \ldots, Z_{t,r}^{(i)} ] \in \mathcal{Z}_t$ denote a vector of true but unobserved/hidden confounders. Let $\mathbf{X}_t^{(i)} = [ X_{t,1}^{(i)}, X_{t,2}^{(i)}, \ldots, X_{t,p}^{(i)} ] \in \mathcal{X}_t$ denote a vector of observed covariates, which is a collection of noisy proxies of the true confounders (\ie, each $X_{t,j}^{(i)}$ is a noisy realization of arbitrary function of $\mathbf{Z}_t^{(i)}$). Further, let $\mathbf{A}_t^{(i)} = [ A_{t,1}^{(i)}, A_{t,2}^{(i)}, \ldots, A_{t,k}^{(i)} ] \in \mathcal{A}_t $ denote a vector of assigned binary treatments and $\mathbf{Y}_{t+1}^{(i)} \in \mathcal{Y}_{t+1}$ the observed one-dimensional outcome. We assume that all static patient covariates, such as gender or genetic information, are part of the observed covariates. We present an overview of the causal structure in \figureref{fig:causal_structure}.


Key to our task is observational data $\mathcal{D}$. The observational data comprises independent patient trajectories from patients $i = 1, \ldots, N$ across time steps $t = 1,\ldots,T$, given by $\mathcal{D} = (\{ \mathbf{x}_t^{(i)}, \mathbf{a}_t^{(i)}, \mathbf{y}_{t+1}^{(i)} \}_{t = 1}^{T} )_{i = 1}^N$. We emphasize that the true confounders, $\mathbf{z}_t^{(i)}$, are missing in $\mathcal{D}$, since they are unobserved, \ie, hidden. Instead, we observe only the noisy proxies, $\mathbf{x}_t^{(i)}$, of the true but hidden confounders. 

We introduce the following notation. Let $\bar{\mathbf{H}}_t = ( \bar{\mathbf{A}}_{t-1}, \bar{\mathbf{X}}_t)$ denote the observed patient history until time $t$, where $\bar{\mathbf{A}}_{t-1} = [ \mathbf{A}_1, \mathbf{A}_2, \ldots, \mathbf{A}_{t-1} ]$ is the history of assigned treatments and $\bar{\mathbf{X}}_t = [ \mathbf{X}_1, \mathbf{X}_2, \ldots, \mathbf{X}_t ]$ is the history of observed covariates (\ie, the history of noisy proxies). We omit the superscript $(i)$ unless it is explicitly required. We use uppercase letters to represent random variables and lowercase letters for realizations of random variables. 

\subsection{Task: ITE Estimation with Noisy Proxies}

We build upon the potential outcomes framework proposed by \citet{Neyman1923} and \citet{Rubin1978}, and extended by \citet{Robins2008} to account for time-varying treatments. Based on this, we denote $\mathbf{Y}_{t + \tau} [\bar{\mathbf{a}}_{(t, t + \tau - 1)}]$ as a potential outcome for a possible treatment assignment $\bar{\mathbf{a}}_{(t, t + \tau - 1)} = [\mathbf{a}_t, \ldots , \mathbf{a}_{t + \tau - 1}]$. If $\bar{\mathbf{a}}_{(t, t + \tau - 1)}$ corresponds to the observed treatment assignment in the data, then the outcome is observed (\ie, the factual outcome). Otherwise, the outcome is unobserved (\ie, the counterfactual outcome). 


Our objective is to estimate the individualized treatment effect (ITE) in the longitudinal setting 
\begin{equation}
\mathbb{E} \left[ \mathbf{Y}_{t + \tau} [\bar{\mathbf{a}}_{(t, t + \tau - 1)}] \mid \bar{\mathbf{H}}_t \right] 
\end{equation}
for $\tau$ steps ahead. We use the causal diagram from \figureref{fig:causal_structure}. In particular, the observed history $\bar{\mathbf{H}}_t$ contains the observed noisy proxies $\bar{\mathbf{X}}_t$, but not the true hidden confounders $\bar{\mathbf{Z}}_t$.

In this paper, we make two standard assumptions for treatment effect estimation over time \citep{Robins2008}: (i)~\emph{(Consistency.)} If $\mathbf{A}_t = \mathbf{a}_t$, then the potential outcome for treatment assignment $\mathbf{a}_t$ is the same as the observed outcome, \ie, $\mathbf{Y}_{t + 1}[\mathbf{a}_t] = \mathbf{Y}_{t + 1}$, and (ii)~\emph{(Positivity.)} If $\mathbb{P}(\bar{\mathbf{A}}_{t-1} = \bar{\mathbf{a}}_{t-1},  \bar{\mathbf{X}}_{t} = \bar{\mathbf{x}}_{t}) \neq 0$, then $\mathbb{P}(\mathbf{A}_t = \mathbf{a}_t\mid \bar{\mathbf{A}}_{t-1} = \bar{\mathbf{a}}_{t-1},  \bar{\mathbf{X}}_{t} = \bar{\mathbf{x}}_{t}) > 0, \, \forall \mathbf{a}_t \in \mathcal{A}_t$. 

Based on the observational data, we can only estimate $\mathbb{E} \left[ \mathbf{Y}_{t + \tau} \mid \bar{\mathbf{a}}_{(t, t + \tau - 1)}, \bar{\mathbf{H}}_t \right]$, but not $\mathbb{E} \left[ \mathbf{Y}_{t + \tau} [\bar{\mathbf{a}}_{(t, t + \tau - 1)}] \mid \bar{\mathbf{H}}_t \right]$. For this reason, many existing methods assume that these quantities are equal by assuming sequential strong ignorability given the \emph{observed} history $\bar{\mathbf{H}}_t$, \ie,
\begin{equation}
\label{ass:seq_stron_ignorability}
\mathbf{Y}_{t + 1} [\mathbf{a}_t] \perp \!\!\! \perp \mathbf{A}_t \mid \bar{\mathbf{H}}_t,
\end{equation} 
for all $\mathbf{a}_t \in \mathcal{A}_t$ and $t$. This assumption would \emph{only} hold true if the true confounders were included in the observed history $\bar{\mathbf{H}}_t$. In our setting, we observe noisy proxies $\bar{\mathbf{X}}_t$ instead of true confounders $\bar{\mathbf{Z}}_t$ (\ie, $\bar{\mathbf{X}}_t$ are a part of observed history $\bar{\mathbf{H}}_t$, and $\bar{\mathbf{Z}}_t$ are unobserved). Hence, sequential strong ignorability is violated, and tailored methods are needed for ITE estimation in the longitudinal setting with noisy proxies.


In this paper, we address this problem by proposing the \longname, which learns a hidden embedding that reflects the true hidden confounders $\mathbf{Z}_t$ using their observed noisy proxies $\mathbf{X}_t$. We emphasize that our objective is not to learn the exact true hidden confounders. Rather, we are interested in a hidden embedding that reflects the true hidden confounders in a way that it can be used to control for them in the downstream task of ITE estimation (\ie, such that we can control for the true hidden confounders by conditioning on the learned hidden embedding). In other words, we want to learn a hidden embedding for which the sequential strong ignorability holds the same way it holds for the true hidden confounders. Once we learn such hidden embedding, we can control for it and obtain unbiased ITE estimates. 

\section{The Deconfounding Temporal Autoencoder}

Based on the above setting, we now develop the \longname, a method that leverages noisy proxies of true hidden confounders for ITE estimation in the longitudinal setting. We introduce our \DTA as follows: in \sectionref{sec:architecture}, we present the architecture of the \DTA that is used to produce the hidden embedding, and, in \sectionref{sec:loss}, we describe the tailored loss function which ensures that \DTA learns the hidden embedding such that it can be used to control for the true hidden confounders.

\subsection{Architecture of the DTA}
\label{sec:architecture}

The architecture of the \DTA consists of two main components: (1)~an \textbf{encoder}, which maps the observed noisy proxies $\mathbf{x}_t$ onto a hidden embedding $\hat{\mathbf{z}}_{t}$; and (2)~a \textbf{decoder}, which uses the hidden embedding $\hat{\mathbf{z}}_{t}$ to generate predictions for the noisy proxies and for the outcome. These predictions are needed for facilitating the learning in such way that the learned hidden embedding can be used to control for the true hidden confounders. 

\vspace{0.2cm}
\textbf{(1) Encoder.} The encoder takes the observed noisy proxies $\mathbf{x}_t$ as input and, based on it produces a hidden embedding $\hat{\mathbf{z}}_{t}$ as the output. Recall that we are in the longitudinal setting, in which the hidden confounders are time-varying. Owing to this, our implementation for learning the hidden embedding is based on a recurrent neural network (RNN). In particular, we use a long short-term memory (LSTM) network \citep{Hochreiter1997}. The LSTM is effective in handling different lag dependencies in time series while mitigating the vanishing (or exploding) gradient problem. Hence, in the first step, the LSTM layer takes the observed noisy proxies $\mathbf{x}_t$ as input, and outputs hidden states $\mathbf{h}_t$. Following the LSTM layer, the encoder uses a linear fully-connected layer to map the output of the LSTM layer onto a hidden embedding, \ie,
\begin{equation}
\label{eq:embedding}
\hat{\mathbf{z}}_{t} = \mathbf{W}_{hz} \mathbf{h}_t + \mathbf{b}_{z}.
\end{equation}
The output of the encoder, $\hat{\mathbf{z}}_{t}$, represents the hidden embedding which is later used for ITE estimation.

\vspace{0.2cm}
\textbf{(2) Decoder.} The decoder takes the hidden embedding $\hat{\mathbf{z}}_{t}$ as input and, based on it, generates two predictions as output: (i)~it predicts the noisy proxies $\hat{\mathbf{x}}_{t}$, and (ii)~it predicts the outcome $\hat{\mathbf{y}}_{t+1}$. 

\begin{figure}[tbp]
\centerline{\includegraphics[width=6.25cm]{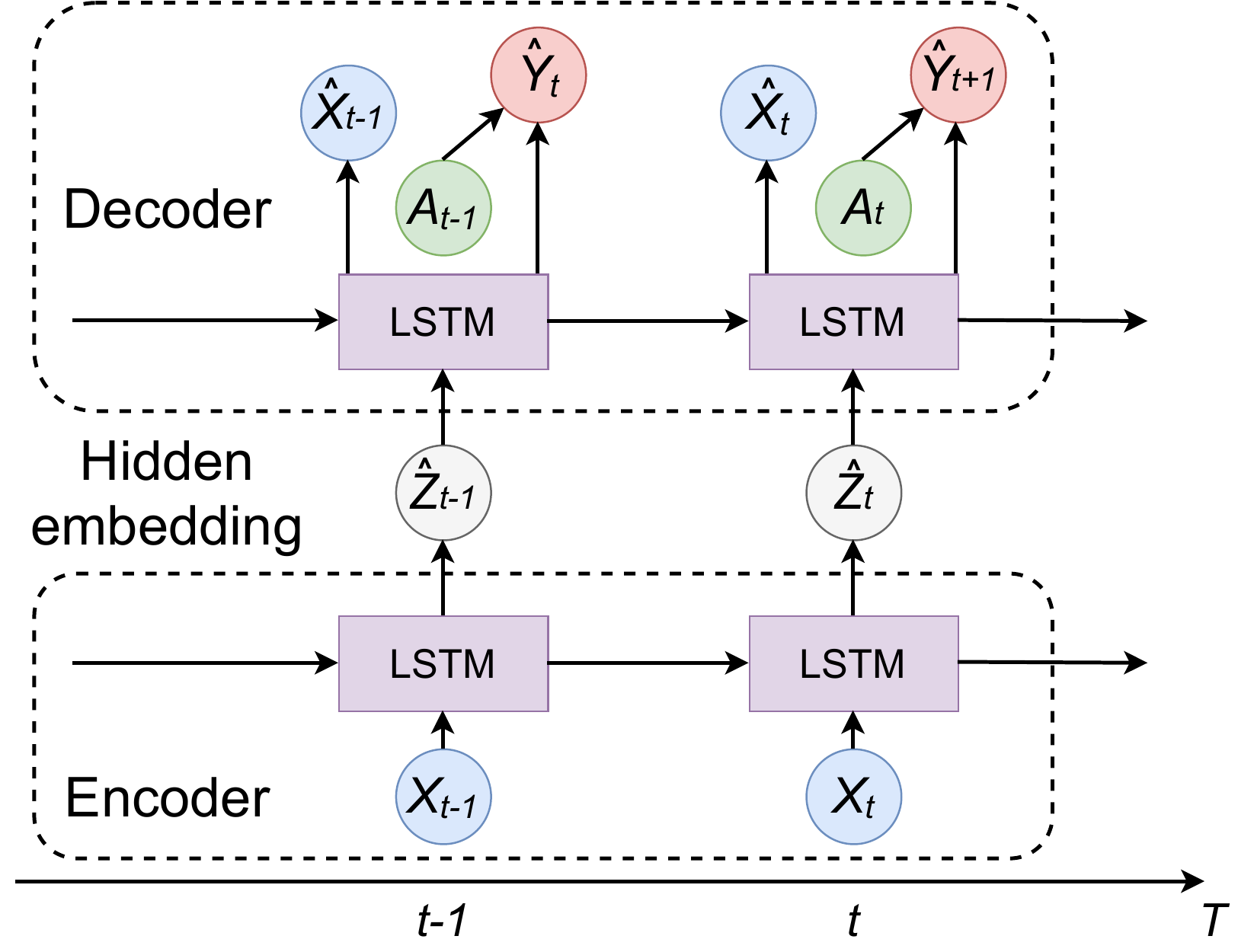}}
\vspace{-.75em}\caption{An overview of the \DTA architecture.}
\label{fig:DTA_architecture}\vspace{-1.25em}
\end{figure}

The first part of the decoder architecture is the same as for the encoder architecture, \ie, we apply an LSTM layer to a given input. However, the input for the LSTM layer of the decoder is the hidden embedding $\hat{\mathbf{z}}_{t}$, and the outputs are the hidden states that we denote as $\mathbf{h'}_t$ to differentiate from the encoder hidden states $\mathbf{h}_t$. The hidden states $\mathbf{h'}_t$ are then used to compute two predictions: (i)~We predict the noisy proxies $\mathbf{x}_t$, \ie,
\begin{equation}
\label{eq:xpred}
\hat{\mathbf{x}}_t = \mathbf{W}_{hx} \mathbf{h'}_t + \mathbf{b}_{x}
\end{equation} 
via a linear fully-connected layer. (ii)~We predict the outcome $\mathbf{y}_{t+1}$ via another linear fully-connected layer, \ie,
\begin{equation}
\label{eq:ypred}
\hat{\mathbf{y}}_{t+1} = \mathbf{W}_{hy} [ \mathbf{h'}_t; \mathbf{a}_t ] + \mathbf{b}_{y} .
\end{equation}
Here, $[\,\cdot;\cdot\,]$ denotes a concatenation. The outcomes are thus predicted using the hidden state $\mathbf{h'}_t$ together with the assigned treatments $\mathbf{a}_t$. We give an overview of the \DTA architecture in \figureref{fig:DTA_architecture}.

Altogether, the \DTA generates a hidden embedding $\hat{\mathbf{z}}_{t}$, and predictions for noisy proxies and the outcome, \ie, $\hat{\mathbf{x}}_{t}$ and $\hat{\mathbf{y}}_{t+1}$, respectively. These outputs are later inserted in the loss function as part of the learning algorithm. 

\subsection{Loss function}
\label{sec:loss}

In \DTA, we aim at leveraging observed noisy proxies to learn a hidden embedding that can be used for unbiased ITE estimation. Hence, our objective is that the hidden embedding $\hat{\mathbf{Z}}_{t}$, \ie, the output of our encoder, is such that it can be used to control for the true hidden confounders $\mathbf{Z}_{t}$ during ITE estimation. Therefore, the hidden embedding $\hat{\mathbf{Z}}_{t}$ should satisfy the following: 
\begin{enumerate}
\item[(1)] $\hat{\mathbf{Z}}_{t}$ should capture the relevant information on true hidden confounders from observed noisy proxies, while reducing noise.
\item[(2)] $\hat{\mathbf{Z}}_{t}$ should allow for good prediction of the potential outcomes. 
\item[(3)] $\hat{\mathbf{Z}}_{t}$ should render the potential outcomes and assigned treatments conditionally independent (\ie, when $\hat{\mathbf{Z}}_{t}$ is included in the patient history, the sequential strong ignorability holds).
\end{enumerate}
We translate the above (1)--(3) into the following loss functions: \textbf{(1)~reconstruction loss}  $\mathcal{L}_x$; \textbf{(2)~outcome prediction loss} $\mathcal{L}_y$; and \textbf{(3)~causal regularization penalty} $\mathcal{R}$. These three loss functions are then later combined into an overall loss function $\mathcal{L}$. In the following, we explain the structure and the purpose for each of the loss components.

\vspace{0.2cm}
\textbf{(1) Reconstruction loss  $\mathcal{L}_x$.} In our given setting, the observed covariates $\mathbf{X}_{t}$ represent noisy proxies of the true hidden confounders $\mathbf{Z}_{t}$. Hence, our hidden embedding $\hat{\mathbf{Z}}_{t}$ (that reflects the true hidden confounders) should capture the relevant information from noisy proxies while reducing noise. We achieve this using the reconstruction loss
\begin{equation}
\label{loss:rec}
\mathcal{L}_x = \frac{1}{N} \frac{1}{T} \sum_{i=1}^N \sum_{t=1}^T \frac{1}{p} \big\| \mathbf{x}_{t}^{(i)} - \hat{\mathbf{x}}_{t}^{(i)} {\big\|}_2^2 ,
\end{equation}
where $N$ is the number of patients, $p$ is the size of observed covariate space, $\| \cdot \|_2$ is the Euclidean norm, $\mathbf{x}_{t}$ are the observed noisy proxies, and $\hat{\mathbf{x}}_{t}$ is the output of the \DTA encoder-decoder architecture. 


The above essentially gives an LSTM autoencoder. This can be seen as follows. Recall that $\hat{\mathbf{x}}_{t}$ is obtained by applying the encoder with LSTM layer on the observed noisy proxies $\mathbf{x}_{t}$ to obtain the hidden embedding $\hat{\mathbf{z}}_{t}$ (see~(\ref{eq:embedding})), followed by applying the decoder with LSTM layer to $\hat{\mathbf{z}}_{t}$ (see~(\ref{eq:xpred})). Therefore, the reconstruction loss effectively creates an LSTM autoencoder for the observed noisy proxies. As such, the objective of the LSTM autoencoder is that the hidden embedding $\hat{\mathbf{Z}}_{t}$ captures the relevant information on the true hidden confounders $\mathbf{Z}_{t}$ from their noisy proxies $\mathbf{X}_{t}$, while reducing noise in the process. 

\vspace{0.2cm}
\textbf{(2) Outcome prediction loss $\mathcal{L}_y$.} Recall that \DTA is not an outcome model itself, \ie, we do not use it for predicting outcomes. Rather, \DTA is a method that learns a hidden embedding that reflects the true hidden confounders and, as such, can be used with any outcome model for ITE estimation. Nevertheless, we still include the outcome prediction loss in our learning algorithm for the following rationale: (i)~the property of the true hidden confounders is that they are predictive of the outcome, and (ii)~we use estimates of potential outcomes later for our causal regularization penalty. 

In \DTA, the outcome prediction loss is given by
\begin{equation}
\label{loss:y}
\mathcal{L}_y = \frac{1}{N} \frac{1}{T} \sum_{i=1}^N \sum_{t=1}^T \big\| \mathbf{y}_{t+1}^{(i)} - \hat{\mathbf{y}}_{t+1}^{(i)} {\big\|}_2^2 ,
\end{equation}
where $N$ is the number of independent patients, $\| \cdot \|_2$ is the Euclidean norm, $\mathbf{y}_{t+1}$ is the observed outcome, and $\hat{\mathbf{y}}_{t+1}$ is the outcome prediction from the \DTA encoder-decoder architecture. 


We now compare the outcome prediction loss against our underlying rationale from (i) and (ii). For (i), the reasoning why our outcome prediction loss is sufficient is straightforward. For (ii), we need to elaborate further on how we obtain the estimates for potential outcomes. Recall that we predict the outcome $\mathbf{y}_{t+1}$ by concatenating the hidden state of the LSTM layer of the decoder with the observed treatment assignment $\mathbf{a}_{t}$ (see~(\ref{eq:ypred})). Hence, even though we only have access to factual outcomes, \ie, observed $\mathbf{y}_{t+1}$, we can use the outcome prediction network to obtain estimates for all potential outcomes by changing the treatment assignment $\mathbf{a}_{t}$ in the prediction. This way, for any treatment assignment $\mathbf{a'}_{t}$, we can obtain an estimate for potential outcome $\hat{\mathbf{y}}_{t+1}[\mathbf{a'}_{t}]$ by using the outcome prediction network. We use these estimates for computing the causal regularization penalty.

\vspace{0.2cm}
\textbf{(3) Causal regularization penalty $\mathcal{R}$.} The causal regularization penalty is one of the main technical novelties of our work. It aims at ensuring that the hidden embedding reflects the true hidden confounders, such that the sequential strong ignorability holds when conditioning on the learned hidden embedding, thus allowing for unbiased ITE estimation. 


The theoretical idea of the causal regulization penalty is to measure the the following Kullback-Leibler (KL) divergence, \ie, 
\begin{equation}
\label{eq:reg}
\sum_{ \mathbf{a'}_t \in \mathcal{A}_t} D_{\mathrm{KL}} \left( p_{(\mathbf{Y}_{t+1}[\mathbf{a'}_t] \, | \, \hat{ \bar{\mathbf{Z}}}_t, \bar{\mathbf{A}}_{t-1} \, )} \, \Big \| \, p_{(\mathbf{Y}_{t+1}[\mathbf{a'}_t] \, | \, \hat{ \bar{\mathbf{Z}}}_t, \bar{\mathbf{A}}_{t} \,  )}\right),
\end{equation}
Hence, the causal regularization penalty equals the sum of KL divergence between the conditional distribution of a potential outcome $\mathbf{Y}_{t+1}[\mathbf{a'}_t]$ given the hidden embedding history $\hat{ \bar{\mathbf{Z}}}_t$ and treatment assignment history $\bar{\mathbf{A}}_{t-1}$ until time $t-1$, and the conditional distribution of a potential outcome $\mathbf{Y}_{t+1}[\mathbf{a'}_t]$ given the hidden embedding history $\hat{ \bar{\mathbf{Z}}}_t$ and treatment assignment history $\bar{\mathbf{A}}_{t}$ until time $t$, for all possible potential outcomes, \ie, $\forall \mathbf{a'}_t \in \mathcal{A}_t$.

We formalize the causal regularization penalty in such way that minimizing it forces the sequential strong ignorability to hold with respect to the hidden embedding $\hat{ \bar{\mathbf{Z}}}_t$ that we learn. We achieve this by using the KL divergence as a distribution distance metric which equals zero only when the two distributions are equal, and it is otherwise positive. Hence, for the two conditional distributions above, the KL divergence equals zero only if the potential outcome $\mathbf{Y}_{t+1}[\mathbf{a'}_t]$ is independent of $\mathbf{A}_t$ given the hidden embedding history $\hat{ \bar{\mathbf{Z}}}_t$ and treatment history $\bar{\mathbf{A}}_{t-1}$ until time $t-1$. On the contrary, the larger the conditional dependence between $\mathbf{Y}_{t+1}[\mathbf{a'}_t]$ and $\mathbf{A}_t$, the larger is the expected difference between the distributions $p(\mathbf{Y}_{t+1}[\mathbf{a'}_t] \, | \, \hat{ \bar{\mathbf{Z}}}_t, \bar{\mathbf{A}}_{t-1} \, )$ and $p(\mathbf{Y}_{t+1}[\mathbf{a'}_t] \, | \, \hat{ \bar{\mathbf{Z}}}_t, \bar{\mathbf{A}}_{t} \,  )$ and , thus, the larger is the KL divergence. Therefore, minimizing the causal regularization penalty forces the potential outcomes and assigned treatment to be conditionally independent given the hidden embedding history $\hat{ \bar{\mathbf{Z}}}_t$ and treatment history $\bar{\mathbf{A}}_{t-1}$ until time $t-1$. In other words, the causal regularization penalty enforces the sequential strong ignorability when we condition on the hidden embedding history $\hat{ \bar{\mathbf{Z}}}_t$ and, hence, allows for unbiased ITE estimation. This way, we ensure that the hidden embedding $\hat{ \bar{\mathbf{Z}}}_t$ reflects the true hidden confounders $ \bar{\mathbf{Z}}_t$, such that we can control for them during ITE estimation. 

We now present how we empirically compute the causal regularization penalty used in the overall loss function, since the equation (\ref{eq:reg}) above is a theoretical construct that needs to be estimated using data. First, we assume that the conditional distributions $p(\mathbf{Y}_{t+1}[\mathbf{a'}_t] \, | \, \hat{ \bar{\mathbf{Z}}}_t, \bar{\mathbf{A}}_{t-1} \, )$ and $p(\mathbf{Y}_{t+1}[\mathbf{a'}_t] \, | \, \hat{ \bar{\mathbf{Z}}}_t, \bar{\mathbf{A}}_{t} \,  )$ are Gaussian. Gaussian distribution is a standard approximation used for continuous variables. Further, the KL divergence for two Gaussian distributions $p$ and $q$ is given by
\begin{equation}
D_{\mathrm{KL}}( p \, \| \, q) = \log\left(\frac{\sigma_q}{\sigma_p}\right) + \frac{\sigma_p^2 + (\mu_p - \mu_q)^2}{2 \sigma_q^2} - \frac{1}{2}.
\end{equation}
Finally, to compute the conditional means and variances for $p(\mathbf{Y}_{t+1}[\mathbf{a'}_t] \, | \, \hat{ \bar{\mathbf{Z}}}_t, \bar{\mathbf{A}}_{t-1} \, )$ and $p(\mathbf{Y}_{t+1}[\mathbf{a'}_t] \, | \, \hat{ \bar{\mathbf{Z}}}_t, \bar{\mathbf{A}}_{t} \,  )$, we use a linear regression model that looks at the previous step in the history as an efficient approximation. For this, we use the data on potential outcomes $\hat{\mathbf{y}}_{t+1}[\mathbf{a'}_{t}]$ obtained from the outcome prediction network, hidden embedding $\hat{\mathbf{z}}_t, \hat{\mathbf{z}}_{t-1}$ obtained from the encoder, and observed treatment assignments $\mathbf{a}_t, \mathbf{a}_{t-1}$. Thus, we empirically compute the causal regularization penalty via
\begin{equation}
\label{loss:reg}
\mathcal{R} = \frac{1}{T} \frac{1}{2^k} \sum_{t=1}^{T} \sum_{\forall \mathbf{a'}_t \in \mathcal{A}_t} \hat{D}_{KL_t}^{(\mathbf{a'}_t)}
\end{equation}
with
\begin{equation*}
\hat{D}_{\mathrm{KL}_t}^{(\mathbf{a'}_t)} = \log\left(\frac{\hat{\sigma}_{2,t}}{\hat{\sigma}_{1,t}}\right) + \frac{\hat{\sigma}_{1,t}^2 + \frac{1}{N}\sum_{i=1}^{N}(\hat{\mu}_{1,t}^{(i)} - \hat{\mu}_{2,t}^{(i)})^2}{2 \hat{\sigma}_{2,t}^2} - \frac{1}{2},
\end{equation*}
where $2^k$ is the number of treatment assignment combinations at time $t$, and $\hat{\mu}_{1,t}, \hat{\sigma}_{1,t}$ and $\hat{\mu}_{2,t}, \hat{\sigma}_{2,t}$ are conditional mean and conditional variance estimates from linear regression of $\mathbf{Y}_{t+1}[\mathbf{a'}_t]$ on $\hat{\mathbf{Z}}_t, \hat{\mathbf{Z}}_{t-1}, \mathbf{A}_t, \mathbf{A}_{t-1}$ and on $\hat{\mathbf{Z}}_t, \hat{\mathbf{Z}}_{t-1}, \mathbf{A}_{t-1}$, respectively. 

\textbf{Overall loss $\mathcal{L}$.} We now combine the above loss functions (1)--(3) into an overall loss 
\begin{equation}
\label{eq:obj}
\mathcal{L} =  \mathcal{L}_x + \theta \, \mathcal{L}_y + \alpha \, \mathcal{R},
\end{equation}
where $\mathcal{L}_x$ is the reconstruction loss for noisy proxies, $\mathcal{L}_y $ is the outcome prediction loss, and $\mathcal{R}$ is the causal regularization penalty. The variables $\theta$ and $\alpha$ are hyperparameters that control the relative trade-off between the three loss components.

Altogether, the overall loss function combines the three objectives for the hidden embedding: (1)~capturing the relevant information from noisy proxies while reducing noise; (2)~being predictive of the outcome, and (3)~rendering estimated potential outcomes and treatment assignment conditionally independent. Hence, the overall objective is to leverage the information from observed noisy proxies to learn a hidden embedding for which the sequential strong ignorability is satisfied, thus allowing for unbiased ITE estimation. We provide the learning algorithm and implementation details for the \DTA in \appendixref{app:algo}.

\subsection{\DTA for ITE estimation}

Our \DTA generates a hidden embedding that can be used to control for the true hidden confounders in the downstream task of ITE estimation. As such, the \DTA is used in conjunction with an outcome model for ITE estimation in the longitudinal setting. By conditioning on the hidden embedding, outcome models produce unbiased estimates of ITE. On the contrary, if outcome models are used without the \DTA (\ie, if they condition on the observed noisy proxies instead of on the hidden embedding), they produce biased ITE estimates. Therefore, we later show that our \DTA improves the performance of state-of-the-art outcome models for ITE estimation in the setting where there are hidden confounders with observed noisy proxies.  

\textbf{Outcome models.} In order to obtain outcome predictions, \DTA must be used in conjunction with an outcome model that is designed for ITE estimation in the longitudinal setting. We combine \DTA with two established outcome models (but emphasize that users can also choose any outcome models if desired): (i)~\emph{Marginal structural models (MSMs)} - an established method for longitudinal ITE estimation in epidemiology literature \citep{Robins2000}, and (ii)~\emph{Recurrent marginal structural networks (\mbox{RMSNs})} - a state-of-the-art machine learning method for ITE estimation in the longitudinal setting \citep{Lim2018}. When we use the \DTA in conjunction with the outcome models, then it is effectively an ITE estimation method.

\section{Experiments}
\label{sec:experiments}

In this section, we demonstrate the effectiveness of the \DTA in experiments using both (i)~synthetic and (ii)~real-world data: (i)~Synthetic data allow us to simulate data in which we control the amount of hidden confounding and where we can thus compare \DTA against an oracle. (ii)~Real-world data allow us to demonstrate potential benefits in a practical application. Here, we use a medical setup, specifically from intensive care units, which is known to be prone to many hidden confounders \citep{Gottesman2019}. Across both, we show that the \DTA reduces bias in ITE estimation when used in conjunction with an outcome model. For this, we compare outcome predictions when conditioning on observed covariates (\ie, noisy proxies of true hidden confounders) versus predictions when conditioning on the hidden embedding learned by the \DTA.

\subsection{Experiments with synthetic data}

\textbf{Simulated data.} In line with our discussed setting, we simulate the observational data on patient trajectories $\mathcal{D} = (\{ \mathbf{x}_t^{(i)}, \mathbf{a}_t^{(i)}, \mathbf{y}_{t+1}^{(i)} \}_{t = 1}^{T} )_{i = 1}^N$, together with hidden confounders $ ( \{ \mathbf{z}_t^{(i)} \}_{t=1}^T )_{i=1}^N$, similar to \citet{Bica2020tsd}. We simulate datasets consisting of $5000$ patient trajectories with $T=30$ time steps, $r=5$ hidden confounders, $p=20$ observed covariates (\ie, noisy proxies of hidden confounders), and $k=2$ different treatments. We vary the confounding parameter $\gamma \in \{0, 0.2, 0.4, 0.6, 0.8 \}$ to produce datasets with a varying amount of hidden confounding. Further, we also simulate counterfactual outcomes for a random treatment assignment, which are later used to evaluate our method against an oracle (note that such an oracle would not be available in a real-world setting). Simulation details are provided in \appendixref{app:data}.
\begin{figure*}[ht]
\centering
\centerline{\includegraphics[width=13cm, height=5cm]{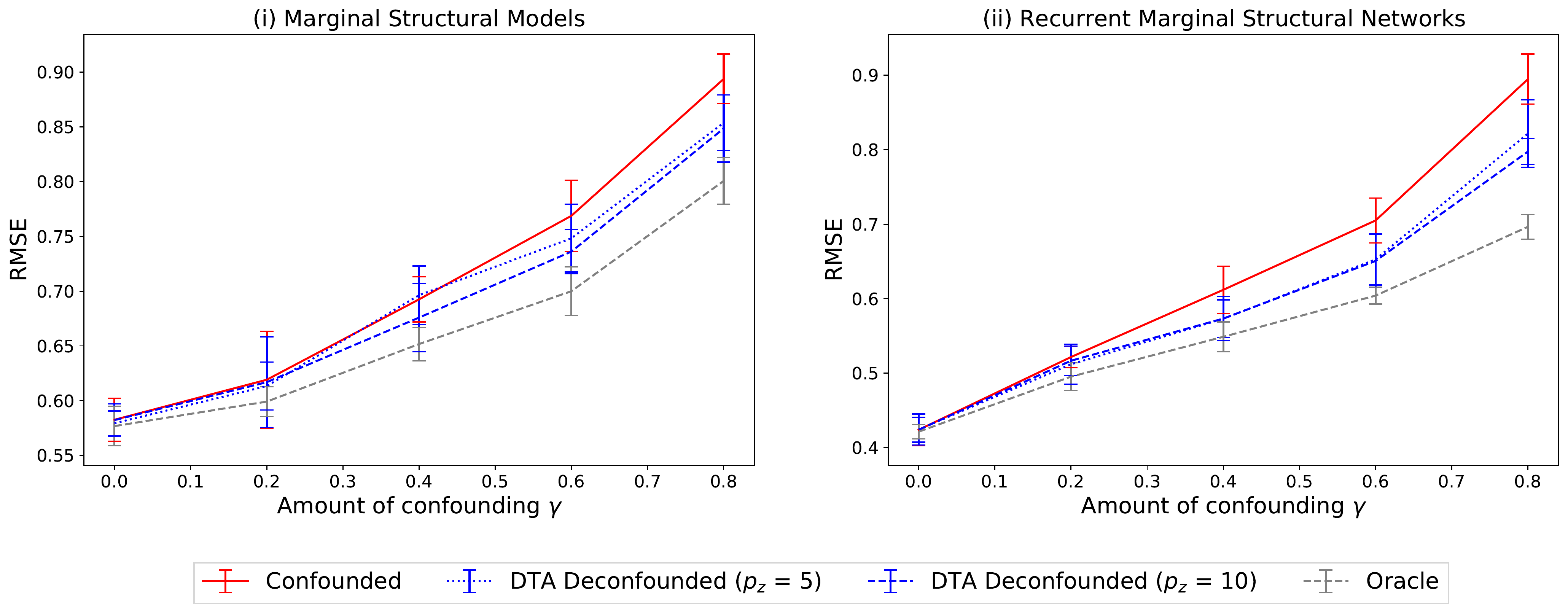}}
\vspace{-1.25em}\caption{The results from experiments on synthetic data.}
\label{fig:results}
\vspace{-1.5em}
\end{figure*}

\textbf{Methods/baselines.} Our \DTA is evaluated by examining how well can it control for the hidden confounders. Hence, suitable baselines would be methods for ITE estimation in the longitudinal setting that aim to control for the hidden confounders. However, existing methods for this purpose cannot be applied in our longitudinal setting with noisy proxies (see \sectionref{sec:lit}). Nonetheless, one way to evaluate the \DTA is by comparing the performance of outcome models with \DTA versus without \DTA. Hence, viable baselines are outcome models for ITE estimation in the longitudinal setting, since any such outcome model can be used for comparison. We emphasize that we examine relative improvements by adding \DTA to an outcome model, and hence, the choice of the outcome model is not important for evaluating \DTA performance. We have chosen MSMs \citep{Robins2000} and RMSNs \citep{Lim2018} as established outcome models from epidemiology and machine learning literature, respectively.   

We compare the following methods (implementation details are provided in \appendixref{app:exp}):
\begin{enumerate}
\item \textbf{Confounded (MSM)} and \textbf{Confounded (RMSN)} represent our main baselines as two state-of-the-art outcome models for ITE estimation in the longitudinal setting. These have access to identical data as our \DTA model: the baselines consider time-varying covariates but without accounting for that these are noisy proxies of the true hidden confounders. Formally, they use the observed noisy proxies to estimate $ \mathbb{E} \left[ \mathbf{Y}_{t + \tau} \mid \bar{\mathbf{a}}_{(t, t + \tau - 1)},\bar{\mathbf{A}}_{t-1}, \bar{\mathbf{X}}_t \right]$. We know that these ITE estimates are biased when hidden confounders are present because the noisy proxies $\bar{\mathbf{X}}_t$ do not satisfy the sequential strong ignorability (see~(\ref{ass:seq_stron_ignorability})).
\item  \textbf{\DTA Deconfounded (MSM)} and \textbf{\DTA Deconfounded (RMSN)} represent our proposed method together with two separate outcome models. These use the hidden embedding $\hat{\bar{\mathbf{Z}}}_t$ learned by the \DTA instead of noisy proxies $\bar{\mathbf{X}}_t$ and, thus, estimate $ \mathbb{E} \left[ \mathbf{Y}_{t + \tau} \mid \bar{\mathbf{a}}_{(t, t + \tau - 1)},\bar{\mathbf{A}}_{t-1}, \hat{\bar{\mathbf{Z}}}_t \right]$. In the simulation study, we further compare two variants with two different sizes of the hidden embedding, namely $p_z = p/4 = 5$ and $p_z = p/2 = 10$. This is to show that the results are robust to different hyperparameters.
\item \textbf{Oracle (MSM)} and \textbf{Oracle (RMSN)} are for benchmarking purposes only. These use knowledge of the true hidden confounders $\bar{\mathbf{Z}}_t$ to estimate $ \mathbb{E} \left[ \mathbf{Y}_{t + \tau} \mid \bar{\mathbf{a}}_{(t, t + \tau - 1)},\bar{\mathbf{A}}_{t-1}, \bar{\mathbf{Z}}_t \right]$. Such knowledge is only available in simulation studies and not in real-world applications. 
\end{enumerate}


\textbf{Results.} \figureref{fig:results} evaluates the different methods (left: MSMs; and right: RMSNs). Here, performance is reported based on the root mean squared error (RMSE) with respect to five-step-ahead prediction of counterfactual outcomes. Most importantly, our proposed \DTA Deconfounded (\textcolor{blue}{in blue}) is superior to the state-of-the-art baselines in form of the plain outcome models named Confounded (\textcolor{red}{in red}). The performance improvement from our \DTA Deconfounded methods becomes larger with increasing amount of confounding $\gamma$. These patterns remain robust across both outcome models (MSMs and RMSNs). Note that both the Confounded baselines and our \DTA Deconfounded method have access to the identical data, and, hence, the performance gain must be attributed to more effective handling of noisy proxies. 

We make further observations. First, we compare two variants our \DTA Deconfounded, in which we vary in the size of the hidden embedding $p_z$. Here, we barely see differences. Evidently, the performance of \DTA Deconfounded remains fairly robust to changes in this hyperparameter. Second, we see that predictions from \DTA Deconfounded come close to the {Oracle} (\textcolor{gray}{in gray}), \ie, the predictions that have access to the true hidden confounders. Altogether, the results demonstrate that conditioning on the hidden embedding learned by the \DTA offers superior outcome predictions compared to conditioning on observed noisy proxies, and thus, reduces the bias in ITE estimation.

\subsection{Experiments with real-world data}


\textbf{Real-world data.} We now study the performance of the \DTA on a real-world dataset where we do not have the knowledge of the true hidden confounders. Here, we draw upon a common benchmarking dataset, namely MIMIC-III \citep{Johnson2016}. MIMIC-III consists of patient trajectories from intensive case units. In our experiment, we use an existing pre-processing pipeline provided by \citet{Wang2020}. We extracted 2313 patient trajectories with 30 time steps each. For each patient, we extracted 25 time-varying covariates that include vital signs and lab tests measured over time (see codes for details). We further included two treatments: vasopressors and mechanical ventilation. The data is aggregated by hour (\ie, we use hourly means for the time-varying covariates and treatments). We then estimated the effect of vasopressors and mechanical ventilation on two outcome variables: (diastolic) blood pressure and oxygen saturation. 

\textbf{Methods/baselines.} For this real-world dataset, we must expect that hidden confounders are present. MIMIC-III includes several measurements of vital signs and lab tests, yet these are -- by no means -- complete and thus cannot fully capture a patient's health status. Moreover, MIMIC-III is a real-world observational dataset, and, because of that, we no longer have access to the true hidden confounders (nor have ways to infer them with certainty). As a consequence, estimates of true ITE are also not available, and, hence, we can only compare the outcome prediction performance when conditioning on the hidden embedding learned by the \DTA (labeled {\DTA Deconfounded}) versus conditioning on observed covariates (labeled {Confounded}).  The latter corresponds again to the plain outcome models from the state-of-the-art, namely MSMs and RMSNs. The implementation details are provided in \appendixref{app:exp}.

\textbf{Results.} \tableref{tab:results} shows the results. By comparing the RMSE, we see that the \DTA Deconfounded substantially improves the outcome prediction. This finding is consistent for both outcome variables (\ie, blood pressure and oxygen saturation) and for both outcome models (\ie, MSMs and RMSNs). For example, the prediction error of the MSM reduces, on average, by 14.7\,\% for blood pressure and by 26.9\,\% for oxygen saturation. The results show that the \DTA Deconfounded produces superior outcome predictions compared to the baselines on a real-world medical dataset.

\begin{table}[ht]
\caption{Results for real-world data experiments.}\vspace{-.75em}
\begin{center}
\resizebox{\columnwidth}{!}{
\begin{tabular}{l c c c c}
\hline \hline
\textbf{Outcome} &  \multicolumn{4}{c}{\textbf{RMSE (Mean $\pm$ SD)}} \\
\cline{2-5}
& \multicolumn{2}{c}{\textbf{MSM}} & \multicolumn{2}{c}{\textbf{RMSN}} \\
\cline{2-3} \cline{4-5}
 & Conf. & \DTA Deconf. & Conf. & \DTA Deconf. \\
\hline
Blo. pre.  & 10.87 $\pm$ 0.00 & \textbf{9.27} $\pm$ \textbf{0.09}  & 12.12 $\pm$ 0.12 & \textbf{11.84} $\pm$ \textbf{0.02} \\
\hline
Oxy. sat. & 4.89 $\pm$ 0.00 & \textbf{3.57} $\pm$ \textbf{0.02} & 6.43 $\pm$ 0.17 & \textbf{5.74} $\pm$ \textbf{0.19} \\
\hline \hline
\multicolumn{4}{l}{Lower = better; best in bold}
\end{tabular}
}
\label{tab:results}
\end{center}\vspace{-2.75em}
\end{table}

\section{Conclusion}

In this paper, we develop the \longname (\DTA), a novel method that leverages observed noisy proxies of confounders for ITE estimation in the longitudinal setting. Formally, our \DTA combines (i)~an LSTM autoencoder with (ii)~a causal regularization penalty. The LSTM autoencoder learns a hidden embedding that captures relevant information on the true hidden confounders from observed noisy proxies while reducing potential noise. The causal regularization penalty then forces the potential outcomes and treatment assignment to be conditionally independent given the hidden embedding. The output of our \DTA is a hidden embedding that can be used to control for hidden confounders during ITE estimation. As such, \DTA is used in conjunction with an outcome model for ITE estimation in the longitudinal setting to obtain unbiased ITE estimates. We demonstrate the effectiveness of our \DTA across several experiments using both synthetic and real-world medical data. These confirm that our \DTA provides superior ITE estimates. As such, our \DTA has implications for practice, where it allows for improved decision support. For instance, in medicine, our \DTA enables a better personalization of treatment recommendations to individual patients with the prospect of improved treatment effectiveness.

\bibliography{refs}

\appendix

\section{Additional related work}\label{app:relwork}

We further review literature in the following domains: (i) additional methods for longitudinal treatment effect estimation, (ii) longitudinal treatment effect estimation and personalized medicine.

(i)~In \sectionref{sec:lit}, we focused our literature review on established and state-of-the-art methods for longitudinal ITE estimation. Here, we expand this review to a few additional methods. In statistics literature, there is increasing use of targeted maximum likelihood estimation (TMLE) for causal inference. TMLE is a semi-parametric estimation method which is attractive due to its desirable asymptotic properties such as being doubly robust, and allowing for the use of flexible machine learning methods for nuisance parameters. \citet{Schomaker2019} and \citet{Sofrygin2019} use longitudinal TMLE to estimate treatment effects over time in a medical setup. In the domain of longitudinal deconfounding, we further refer to the works of \citet{Liu2020} and \citet{Ma2021}. \citet{Liu2020} propose an outcome model for longitudinal ITE estimation that aims to deal with hidden confounders within outcome prediction framework. However, their method differs from our setting since they consider observed covariates as true confounders and not as noisy proxies of hidden confounders. Similarly, the method proposed by \citet{Ma2021} does not fit our setting either, since their method is tailored for dynamic-networked observational data where confounder representation is learned via graph convolutional networks. 

(ii)~Given the increasing availability of medical data in the form of EHRs, estimating ITE from observational data is becoming crucial for personalized decision-making in medicine \citep{Alaa2017, Lim2018}. For instance, the work on longitudinal ITE estimation by \citet{Lim2018} and \citet{Bica2020crn} is motivated by cancer treatment planning. In this example, accurate ITE estimation would allow for prescribing optimal treatment combination and timing of radiation and chemotherapy based on patient's individual history. Given that medical data is prone to several biases \citep{hattgeneralizing2021} such as hidden confounders \citep{Gottesman2019}, and that EHRs potentially contain noisy proxies thereof, our \DTA can be used as powerful tool to improve individualized decision-making in medicine. Besides treatment effect estimation, there are also important works utilizing machine learning for risk scoring, \ie, predicting patient outcomes based on patient history, which can be important for timing of treatment interventions on individual level \citep{Allam2021, Ozyurt2021}.   

\section{\DTA algorithm}\label{app:algo}

The learning algorithm for the \DTA is given in \algorithmref{alg:DTA}. We implemented the \DTA in PyTorch. The training is done in batches for 200 epochs. We use Adam for optimization \citep{Kingma2015}. The hyperparameters are: the size of the hidden embedding $p_z$, batch size, learning rate, dropout rate on LSTM layers, and overall loss function parameters $\theta$ and $\alpha$. The remaining hyperparameters are standard to LSTMs. The hyperparameters are later tuned via cross-validation (see \appendixref{app:exp}). 

\begin{algorithm2e}
\caption{Learning algorithm for the \DTA}
\label{alg:DTA}
\KwIn{$\{ \mathbf{x}_{t}^{(i)} \}_{t=1}^{T}$, \ie, observed noisy proxies for patient $i$}
\KwOut{$\{ \hat{\mathbf{z}}_{t}^{(i)} \}_{t=1}^{T}$, \ie, hidden embedding for patient $i$}
\begin{algorithmic}[1]
\STATE Randomly initialize the parameters of the \DTA architecture: $\mathbf{W}_{hz}$, $\mathbf{W}_{hx}$, $\mathbf{W}_{hy}$, $\mathbf{b}_{z}$, $\mathbf{b}_{x}$, $\mathbf{b}_{y}$ and parameters of the LSTM architecture.
\STATE Compute $\{ \hat{\mathbf{z}}_{t}^{(i)} \}_{t=1}^{T}$ using (\ref{eq:embedding}).
\STATE Compute $\{ \hat{\mathbf{x}}_{t}^{(i)} \}_{t=1}^{T}$ using (\ref{eq:xpred}).
\STATE \textbf{for} $\mathbf{a'}_t \in \mathcal{A}_t$ \textbf{do}:
\STATE \quad Compute $\{ \hat{\mathbf{y}}_{t+1}^{(i)}[\mathbf{a'}_{t}] \}_{t=1}^{T}$ using (\ref{eq:ypred}).
\STATE Compute the reconstruction loss $\mathcal{L}_x$ using (\ref{loss:rec}).
\STATE Compute the outcome prediction loss $\mathcal{L}_y$ using (\ref{loss:y}).
\STATE Compute the causal regularization penalty $\mathcal{R}$ using (\ref{loss:reg}).
\STATE Compute the overall loss $\mathcal{L}$ using (\ref{eq:obj}).
\STATE Update the parameters according to the gradient of the overall loss function.
\end{algorithmic}
\end{algorithm2e}

\section{Synthetic data}\label{app:data}

The following simulation study enables us to test our method in a controlled environment where we can vary the amount of hidden confounding. At each time step $t$, we simulate $r$ hidden time-varying confounders
\begin{equation}
Z_{t,j} = \frac{1}{h} \sum_{i=1}^h \Big(\lambda_{i,j} \, Z_{t-i,j} + \sum_{l=1}^k \omega_{i,l} \, A_{t-i,l} \Big) + \eta_t ,
\end{equation}
where $j=1, \ldots, r$, $\lambda_{i,j} \sim N(0,0.5^2)$, $l=1,\ldots,k$, $\omega_{i,l} \sim N(1-(i/h),(1/h)^2)$, and $\eta_t\sim N(0,0.1^2)$. The hidden confounders thus follow $h$-order autoregressive process, and are affected by previous treatment assignments. Further, we simulate $p$ observed covariates
\begin{equation}
X_{t,m} = \Big(\sum_{j=1}^r \beta_{j,m} \, Z_{t,j}\Big) + \epsilon_t ,
\end{equation}
where $m=1,\ldots,p$, $\beta_{j,m} \sim N(0,1)$, and $\epsilon_t \sim N(0,5^2)$. This way, the observed covariates $\mathbf{X}_t$ are noisy proxies of the true hidden confounders $\mathbf{Z}_t$, which is our assumed setting. 

We simulate $k$ treatments as follows
\begin{equation}
\pi_{t,l} = \gamma_A \bigg(\frac{1}{r} \sum_{j=1}^r Z_{t,j} \bigg) + (1-\gamma_A) \bigg(\frac{1}{k} \sum_{l=1}^k A_{t-1,l} \bigg) ,
\end{equation}
\begin{equation}
A_{t,l} \, | \, \pi_{t,l} \sim \mathrm{Bernoulli}(\sigma(\pi_{t,l})) ,
\end{equation}
where $l=1,\ldots,k$, $\sigma(\cdot)$ is a sigmoid function, and where parameter $\gamma_A \in [0,1]$ controls the amount of hidden confounding applied to the treatment assignment. For example, $\gamma_A=0$ means that there is no hidden confounding. Finally, we simulate one-dimensional outcome
\begin{equation}
Y_{t+1} = \gamma_Y \bigg(\frac{1}{r} \sum_{j=1}^r Z_{t,j} \bigg) + (1-\gamma_Y) \bigg(\frac{1}{h} \sum_{i=1}^h  \sum_{l=1}^k \omega'_{i,l} \, A_{t-i,l} \bigg),
\end{equation}
where $l=1,\ldots,k$, and $\omega'_{i,l} \sim N(1-(i/h),(1/h)^2)$. Here, similar to $\gamma_A$, the parameter $\gamma_Y \in [0,1]$ controls the amount of hidden confounding applied to the outcome. 

We simulate datasets consisting of $5000$ patient trajectories with $T=30$ time steps, $r=5$ hidden confounders, $p=20$ observed covariates (\ie, noisy proxies of hidden confounders), and $k=2$ different treatments. We use $\gamma = \gamma_A = \gamma_Y$ and vary the confounding parameter $\gamma \in \{0, 0.2, 0.4, 0.6, 0.8 \}$ to produce datasets with a varying amount of hidden confounding. Besides factual outcomes, we also simulate counterfactual outcomes for a random treatment assignment. Counterfactual outcomes are later used to evaluate our method against an oracle (note that such an oracle would not be available in a real-world setting).

\section{\DTA experiments}\label{app:exp}

\textbf{Experiments with synthetic data.} We implement hyperparameter optimization as follows: batch size $\in \{64, 128, 256\}$, learning rate $\in \{0.01, 0.005, 0.001 \}$, dropout rate $\in \{0, 0.1, 0.2, 0.3, 0.4 \}$, and overall loss function parameters $\theta, \alpha \in \{0, 0.5, 1, 2, 5\}$. Hyperparameter optimization is performed via stochastic grid search for 50 iterations. The optimal parameters are chosen via cross-validation.  The outcome models are implemented as in \citet{Robins2008} for MSMs, and as in \citet{Lim2018} for RMSNs. For each amount of confounding $\gamma \in \{0, 0.2, 0.4, 0.6, 0.8\}$, we compute outcome predictions using 10 simulated datasets, where each dataset is split $80/10/10$ for training, validation, and testing, respectively. We evaluate the results for a five-step-ahead counterfactual outcome prediction and report the RMSE.
\\
\\
\noindent
\textbf{Experiments with real-world data.} We use the same outcome models as above (\ie, MSM and RMSN). We tune the same hyperparameters as in the experiments with synthetic data. In addition to them, we consider the size of the hidden embedding $p_z \in \{6, 12, 18\}$ as another hyperparameter. As before, we perform hyperparameter optimization via stochastic grid search and then select the optimal parameters using cross-validation. We split the dataset in $80/10/10$ for training, validation and testing, respectively. We evaluate the results for a five-step-ahead outcome prediction and report the RMSE. The results are averaged over 10 runs.

\end{document}